\documentclass[10pt,twocolumn,letterpaper]{article}

\usepackage{cvpr}
\usepackage{times}
\usepackage{epsfig}
\usepackage{graphicx}
\usepackage{amsmath}
\usepackage{amssymb}

\usepackage{cite}
\usepackage{graphicx}
\usepackage{multirow}
\usepackage{multicol}
\usepackage{subfig}
\usepackage{amsmath}
\usepackage{amsfonts}
\usepackage{romannum}
\usepackage{caption}
\usepackage{tabularx}
\usepackage{tabulary}
\usepackage{array}
\usepackage{tabu}
\usepackage{xcolor}
\usepackage{makecell}

\usepackage[breaklinks=true,bookmarks=false]{hyperref}

\cvprfinalcopy 

\def\cvprPaperID{****} 

\begin{document}

\title{Individual Tooth Detection and Identification from Dental Panoramic X-Ray Images via Point-wise Localization and Distance Regularization}

\author{Minyoung Chung$^1$\quad Jusang Lee$^1$\quad Sanguk Park$^1$\quad Minkyung Lee$^1$\quad Chae Eun Lee$^1$\\\quad Jeongjin Lee$^2$\quad Yeong-Gil Shin$^1$\thanks{Corresponding author (yshin@snu.ac.kr).}
\\\\$^1$Seoul National University, Republic of Korea
\\$^2$Soong-sil University, Republic of Korea
}


\maketitle

\begin{abstract}
   Dental panoramic X-ray imaging is a popular diagnostic method owing to its very small dose of radiation. For an automated computer-aided diagnosis system in dental clinics, automatic detection and identification of individual teeth from panoramic X-ray images are critical prerequisites. In this study, we propose a point-wise tooth localization neural network by introducing a spatial distance regularization loss. The proposed network initially performs center point regression for all the anatomical teeth (i.e., 32 points), which automatically identifies each tooth. A novel distance regularization penalty is employed on the 32 points by considering $L_2$ regularization loss of Laplacian on spatial distances. Subsequently, teeth boxes are individually localized using a cascaded neural network on a patch basis. A multitask offset training is employed on the final output to improve the localization accuracy. Our method successfully localizes not only the existing teeth but also missing teeth; consequently, highly accurate detection and identification are achieved. The experimental results demonstrate that the proposed algorithm outperforms state-of-the-art approaches by increasing the average precision of teeth detection by 15.71\% compared to the best performing method. The accuracy of identification achieved a precision of 0.997 and recall value of 0.972. Moreover, the proposed network does not require any additional identification algorithm owing to the preceding regression of the fixed 32 points regardless of the existence of the teeth.
\end{abstract}

\section{Introduction}

In recent years, many computer-aided diagnosis (CAD) systems have been developed to help clinicians as supplementary tools \cite{van2001computer, doi2007computer, wang2016benchmark}. A significant amount of work burden of clinical experts and the occurrence of misdiagnosis can be reduced by developing CAD systems. Among the imaging protocols, panoramic X-ray imaging is a popular diagnostic method owing to its very small dose of radiation when compared to the cone beam computed tomography \cite{angelopoulos2008comparison}. This method captures the entire oral structure in a two-dimensional (2D) image and provides a noninvasive treatment plan, such as implants and tooth extraction. Moreover, forensic identification can be conducted by analyzing the corresponding individual teeth of the subjects \cite{nomir2007human}. Various applications such as classification \cite{miki2017classification} and segmentation \cite{tuan2016cooperative, tuan2017dental} were developed using dental panoramic X-ray images \cite{wang2016benchmark}. Especially, automated detection and identification of individual teeth are the most demanded algorithms and a critical prerequisite for other applications \cite{chen2019deep}.\par

To localize objects of interest from images, various object detection methods were extensively developed until recently. Initially, classical machine learning-based algorithms were proposed. These classical methods typically employed feature descriptors and trained classifiers to obtain object boxes \cite{dalal2005histograms, zheng2008four}. Recently, deep learning-based approaches have been showing groundbreaking results over the classical methods by exploiting convolutional neural networks (CNNs) \cite{uijlings2013selective, girshick2014rich, he2015spatial, girshick2015fast, ren2015faster, redmon2016you, liu2016ssd, redmon2017yolo9000, redmon2018yolov3, zhou2019objects, law2018cornernet, duan2019centernet}. Modern CNN-based detection methods can be categorized into two primary methods: 1) anchor-based \cite{ren2015faster, redmon2018yolov3} and 2) point-based approaches \cite{zhou2019objects, law2018cornernet, duan2019centernet}. The anchor-based methods employ exhaustive classifications on predefined anchor boxes and typically perform a non-maximum suppression technique to localize each object \cite{ren2015faster}. Conversely, the point-based object detection attempts to regress points to delineate objects such as the center point \cite{duan2019centernet}. Besides the center point, several key points (e.g., left-top and right-bottom corners) are simultaneously regressed for accurate object detection \cite{law2018cornernet, zhou2019bottom}. The latest studies show that the point-based approaches are demonstrating more promising results than the anchor-based methods in terms of accuracy and efficiency \cite{duan2019centernet}.\par

Although CNN-based detection methods are showing groundbreaking results, high accuracy must be guaranteed so that the algorithm can be used as an important auxiliary diagnostic measure in clinical practices. It was reported in \cite{chen2019deep} that an automated deep learning-based algorithm has an impact on teeth detection and identification in dental panoramic images; however, the study showed possible errors in detection, which can result in subsequent identification errors. Thus, the simple adoption of a general CNN-based detection algorithm cannot provide the high standard of accuracy that is required in the clinics. It is critical to build a robust metric that can guarantee accuracy of the algorithm to verify the applicability of the system.\par

In this paper, we propose a CNN-based individual tooth detection and identification algorithm based on direct regression of object points. First, we propose a points regression neural network by employing spatial distance regularization (DR) loss. The proposed network performs center point regression of 32 fixed anatomical teeth, which automatically assigns anatomical identifiers. A novel inter-point DR penalty is employed on the output prediction of 32 points on a neighborhood basis. Subsequently, bounding boxes of an individual tooth is localized in a cascaded fashion. For the final box regressions, a multitask framework is applied for an additional offset vector regression, which is trained to delineate a marginal error vector of a center point. The superiority of the proposed method not only depends on accurate localization but also on automated individual identification of teeth. An additional identification algorithm is not required in the proposed method. The proposed method automatically identifies each tooth by localizing all 32 possible regions of the teeth including missing ones. The experimental results showed that our proposed method outperforms other state-of-the-art methods; moreover, various conditions of test images illustrated the clinical validity of the algorithm. The primary contributions of this work can be summarized as follows:
\begin{itemize}
    \item Integration of a point-based detection method and fixed 32-point regression in a cascaded fashion.
    \item The proposed method does not require any additional classification methods.
    \item Introduction of a DR loss between neighboring teeth to improve the regression.
    \item Multitask training of box parameters and the marginal offset vector of the center point.
\end{itemize}\par

The remainder of this paper is structured as follows. In Section 2, we review related works on object detection and identification methods. Further, we describe our proposed method in Section 3. Section 4 demonstrates the experimental results and Sections 5 and 6 present the discussion and conclusion, respectively.

\section{Object Detection}
In this section, modern CNN-based object detection methods are reviewed. We first present several anchor-based approaches; subsequently, we highlight the state-of-the-art point-based methods that were developed recently.

\subsection{Anchor-based Object Detection}
In the earlier developments of CNN-based object detection, an exhaustive or a selective search of object regions was proposed \cite{uijlings2013selective, girshick2014rich, he2015spatial, girshick2015fast}. The separated two-step training (i.e., extraction of the region-of-interest and the subsequent classification using optional regression) was integrated after the region proposal network (RPN) was proposed \cite{ren2015faster}. The RPN module employed grid-based anchors that were used to perform box regressions \cite{ren2015faster}.\par

Most of the presented methods identified objects as axis-aligned boxes in an image. The candidate boxes of the objects were generated based on predefined anchors with several multiscaled boxes for each anchor \cite{ren2015faster}. An exhaustive classification was performed on all candidate boxes positioned according to predefined anchors \cite{ren2015faster}. To remove multiple overlapping boxes, post-processing localization was required such as non-maximum suppression (NMS) \cite{bodla2017soft}. For real-time applications (e.g., automobile or surveillance vision), one-stage classification and regression networks (i.e., single-shot detectors) were proposed \cite{redmon2016you, liu2016ssd, redmon2017yolo9000, redmon2018yolov3}. An exhaustive classification-based approach, such as faster regions with CNN (R-CNN) \cite{ren2015faster}, was considered to be superior to the single-shot detectors \cite{redmon2016you, liu2016ssd} in terms of accuracy, primarily because of the exhaustive local classifications and post-regression procedure.\par

\begin{figure*}[t!]
    \centering
    \includegraphics[width=\linewidth]{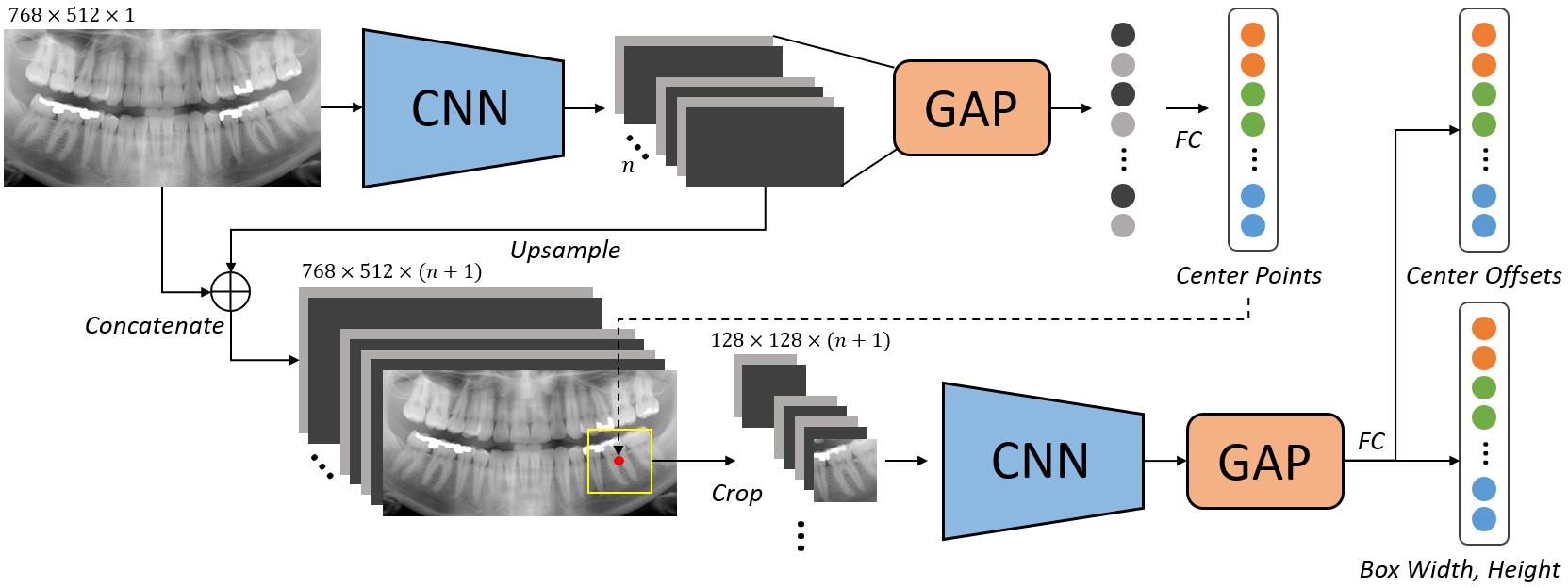}
    \caption{Overall network architecture of the proposed method.}
    \label{fig:network}
\end{figure*}

The anchor-based methods have certain significant drawbacks: 1) the RPN module based on the anchors requires exhaustive classifications for each anchor box \cite{ren2015faster}; 2) the additional classification network requires the memory of an additional GPU; 3) the anchor-based predefined windows exclude small objects in the image. The first exhaustive classification in the RPN module significantly affects the final performance of a detector. That is, the method introduces additional challenges such as a class imbalance problem, NMS processing \cite{bodla2017soft}, and true or negative example mining. 
The method also requires careful tuning of several hyperparameters such as the number of anchors, size of the anchor boxes, and scales.

\subsection{Point-based Object Detection}
More recently, object detection methods that are based on key points were proposed \cite{zhou2019objects, zhou2019bottom, law2018cornernet, duan2019centernet}. The primary concept of these methods is the indication of significant ``points" in the object of interest. In \cite{zhou2019objects}, the authors considered an object as a single point, i.e., the center point of the bounding box of an object. The major advantage of the point-based algorithms is that the architecture does not employ anchor-based extreme classifications. The presented point-based object detection does not require region classification. The bounding boxes of the objects can be obtained by performing regression on orientation and sizes. Point-based methods \cite{zhou2019objects, duan2019centernet} can be considered as anchor-free one-stage approaches \cite{redmon2016you, liu2016ssd, lin2017focal}. The primary differences are that the point-based approaches do not constrain the shapes and anchor positions. The anchor positions are extracted based on key-point detection \cite{newell2017associative, cao2017realtime}; consequently, neither classification nor NMS post-processing \cite{bodla2017soft} is required \cite{zhou2019objects}. The detection of two corner points \cite{law2018cornernet} or five building key points including the center point \cite{zhou2019bottom} was also presented. The center-point-based approaches \cite{zhou2019objects, duan2019centernet} do not require combinatorial grouping of key points after detection; thus, they are simpler and more effective methods when compared to the methods that employ multiple key points \cite{zhou2019bottom}.\par


\section{Methodology}
To accurately localize an individual tooth, we propose a point regression-based object detection method. First, our method performs direct regression of the center points for all possible 32 teeth regardless of the existence of the teeth, which automatically identifies each tooth. Subsequently, teeth boxes are individually localized using a cascaded neural network that utilizes intermediate feature maps for initial predictions and cropped patches from the original image. A simultaneous offset training is employed on the final output based on a multitask framework to improve the accuracy.\par

\subsection{Network Architecture}
As illustrated in Fig. \ref{fig:network}, the proposed network initially estimates the center points of the 32 teeth based on direct regression. We used several backbone networks for the initial 32-point regression, i.e., residual network (ResNet) \cite{he2016deep}, deep layer aggregation (DLA) \cite{yu2018deep}, and stacked hourglass (HG-Stacked) \cite{newell2016stacked}, with a single modification of the final output tensor as a single vector of size 64 for 32 2D points. The output feature maps are pooled using the global average pooling method (GAP) \cite{lin2013network} and subsequently passed through a single fully connected (FC) layer to represent the positions of the 32 points.\par

The final teeth localization is subsequently obtained by a cascading fashion for every patch. The second network utilizes the estimated center positions and intermediate features that were used for the first estimation. The feature maps are upsampled to the original resolution (i.e., $768\times 512$) to concatenate the original input image for further cascaded detection. All 32 patches are cropped to a fixed size of $128\times 128$ corresponding to each center point. For the second CNN, we employed ResNet-18 \cite{he2016deep} for simplicity. The final offsets and box parameters are obtained similar to the first stage through a series of GAP and FC layers.\par

\begin{figure}[t!]
    \centering
    \includegraphics[width=\linewidth]{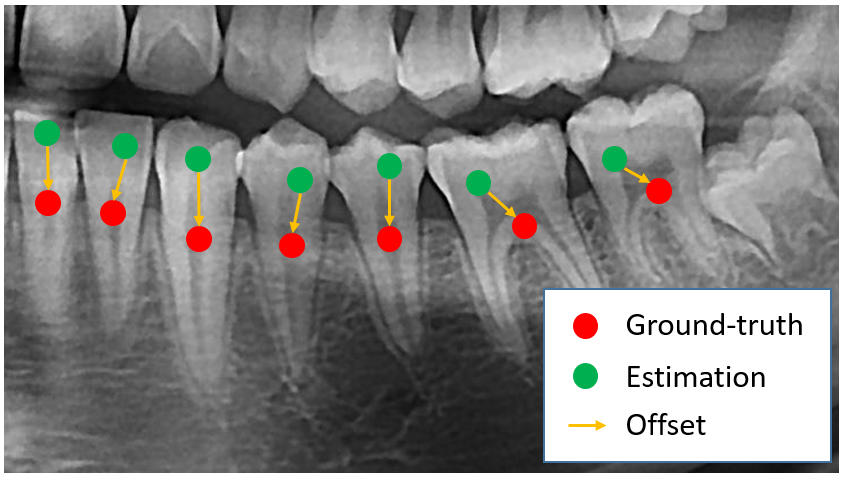}
    \caption{Schematic visualization of offset. Offset is a vector indicating the displacement between the ground-truth center point and the current estimation.}
    \label{fig:offset}
\end{figure}

A multitask training is employed for the final estimation based on two objectives: 1) box parameter regression (i.e., width and height) and 2) center point offset (i.e., marginal vector for center point correction) regression. Figure \ref{fig:offset} shows the ground-truth offsets that are used while training the network. The objective of training offsets is to improve the accuracy of the box position. 

\subsection{Training the Network}

\begin{figure}[t!]
    \centering
    \includegraphics[width=\linewidth]{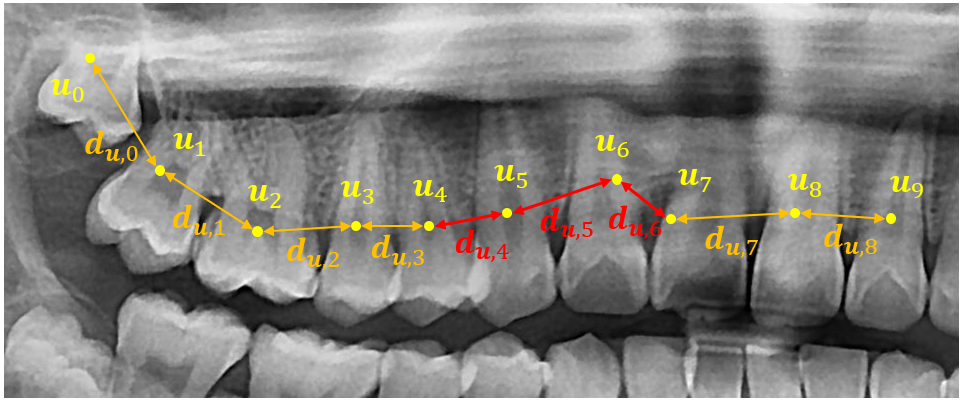}
    \caption{Schematic visualization of distance regularization (DR). The distance vector $\bold{d}$ is defined by spatially ordered distances between neighboring teeth.}
    \label{fig:dr}
\end{figure}

As described in the previous sections, the proposed network is trained based on four different loss metrics: 1) initial regression of center points, 2) DR, 3) offsets of center points (i.e., marginal error vector), and 4) box parameter regression (i.e., width and height). For training the first center point regression, we employed a mean squared error (MSE) function. The loss function can be defined as

\begin{equation}
    L_{cen}=MSE(\bold{x}_{cen}, \bold{y}_{cen}),
\label{eq:center}
\end{equation}
where $\bold{x}_{cen}$ and $\bold{y}_{cen}$ are the estimated vectors of the center points and ground-truth of the center positions (i.e., $\bold{x}_{cen}, \bold{y}_{cen}\in\mathbb{R}^{64}$), respectively. In addition to MSE loss, a novel DR penalty is employed on the output of the initial prediction of 32 points based on $L_2$ regularization loss of Laplacian on spatially aligned inter-distances. If $\bold{x}_{cen}$ is composed of two vectors, then $\bold{x}_{cen}=\{\bold{u}_i, \bold{l}_i | 0\leq i<16, \bold{u}_i\in\mathbb{R}^2, \bold{l}_i\in\mathbb{R}^2\}$, where $\bold{u}_{0..15}$ and $\bold{l}_{0..15}$ indicate the upper and lower teeth position vectors, respectively. Each vector is spatially aligned corresponding to the positions on the ground-truth image (e.g., left to right; Fig. \ref{fig:dr}). We first calculated two neighborhood distance vectors corresponding to the upper and lower position vectors:
\begin{equation}
    \bold{d}_{\bold{u},i}={||\bold{u}_{i+1}-\bold{u}_i||}_2
\label{eq:distance_up}
\end{equation}
and
\begin{equation}
    \bold{d}_{\bold{l},i}={||\bold{l}_{i+1}-\bold{l}_i||}_2,
\label{eq:distance_low}
\end{equation}
where $0\leq i<15$. The distance vectors represent the spatially aligned distances between the neighboring teeth. Finally, we modeled a DR loss corresponding to the distance vectors:
\begin{equation}
    L_{dr}={||\nabla\cdot\nabla \bold{d}_{\bold{u}}||}_2+{||\nabla\cdot\nabla \bold{d}_{\bold{l}}||}_2,
\label{eq:dr}
\end{equation}
where $\nabla$ is a gradient operator. Equation (\ref{eq:dr}) is an $L_2$ regularization of Laplacian on distance vectors. Figure \ref{fig:dr} illustrates $\bold{u}$ and $\bold{d}_\bold{u}$ vectors schematically. The primary underlying principle of the proposed regularization term calculated according to (\ref{eq:dr}) is to smooth the variation of distances between proximate teeth to remove the outlying positions. The red-colored distances in Fig. \ref{fig:dr} demonstrate a high value when calculated using (\ref{eq:dr}), indicating that minimizing the term will regularize the proximate distances for accurate regression.\par

The final offset and box parameter regression is trained similar to (\ref{eq:center}). The offset loss is defined as
\begin{equation}
    L_{off}=MSE(\bold{x}_{off}, \bold{y}_{off}),
\label{eq:offset}
\end{equation}
where $\bold{x}_{off}$ and $\bold{y}_{off}$ are the estimated vector of the final offset and the ground-truth offset at the current iteration, respectively. The ground-truth offset can be calculated by $\bold{y}_{off}=\bold{y}_{cen}-\bold{x}_{cen}$, where $\bold{x}_{cen}$ is the estimated vector of the center points at the current iteration. Similarly, the loss function for bounding box parameters is defined as
\begin{equation}
    L_{box}=MSE(\bold{x}_{box}, \bold{y}_{box}),
\label{eq:box}
\end{equation}
where $\bold{x}_{box}$ and $\bold{y}_{box}$ are the estimated vector of the final box parameters and the ground-truth.\par

The overall loss function is defined by combining all the presented loss functions into a single objective function:
\begin{equation}
    L = L_{cen} + L_{dr} + \alpha L_{off} + \beta L_{box} + \gamma {||W||}_2,
\label{eq:loss}
\end{equation}
where $W$ indicates the weights of the proposed network and $\alpha$, $\beta$, and $\gamma$ are the weighting coefficients. The first two terms in the equation are related to the initial center regression. The third and fourth terms are the offset and box parameter losses, respectively. The final $L_2$ regularization is a global regularization term. We used $\alpha=3, \beta=1.5$, and $\gamma=0.1$ in all the experiments.

\subsection{Training Data}

\begin{figure}[t!]
    \centering
    \subfloat[]{\includegraphics[width=\linewidth]{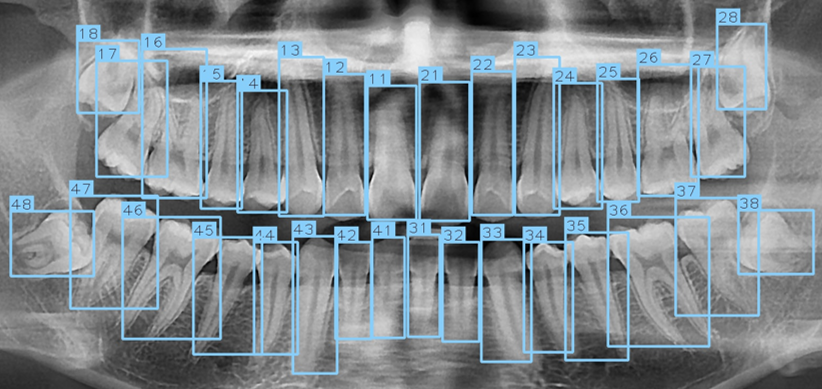}%
    \label{fig:gt_full}}
    \vfil
    \subfloat[]{\includegraphics[width=\linewidth]{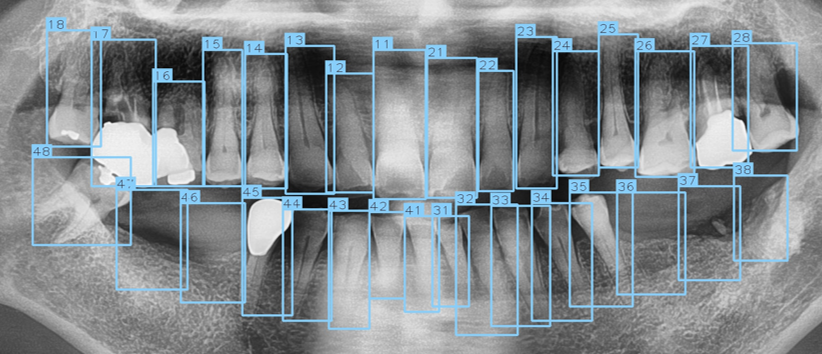}%
    \label{fig:gt_missing}}
    \caption{Visualization of sample annotations: (a) Panoramic image with full teeth (i.e., 32 teeth); (b) Example of several missing teeth. All 32 teeth boxes were annotated regardless of the existence of individual teeth.}
    \label{fig:gt}
\end{figure}

Each image was annotated by clinical experts in the field, as illustrated in Fig. \ref{fig:gt}. All 32 teeth were annotated using axis-aligned bounding boxes with corresponding anatomical identifiers for each tooth. Instead of annotating only the existing teeth in the image, we annotated all 32 teeth boxes, considering even the missing teeth (Fig. \ref{fig:gt_missing}). By enforcing the annotation for all 32 points, the neural network can be trained through direct fixed-point positional regression of all points.\par

All training images were preprocessed using contrast limited adaptive histogram equalization (CLAHE) method \cite{zuiderveld1994contrast}. The primary purpose of employing CLAHE was to minimize the variance of image contrasts among different machines.

\section{Experiments}

\subsection{Data Acquisition and Configuration}
This study was conducted in association with Osstem Implant Co., Ltd. All the datasets were acquired only after the consent of the patients for academic purpose only. The images were sourced from four different machines provided by Osstem Implant, HDX WILL, PointNix, and Genoray. The datasets were obtained from dental clinics for ordinary diagnostic purposes, which indicates that the datasets represent a variety of conditions commonly observed in clinics. In the dataset, the width and height of all panoramic X-ray images ranged from 2093 to 3432 pixels and 1012 to 1504 pixels, respectively. The x- and y-axis spacing in the image ranged from 0.07 to 0.10mm.\par

A total of 818 images were collected for training, validation, and testing. We used 574 images for training, 162 images for validation, and 82 images for testing. The quantitative evaluations for all the experimental results were conducted on 82 test images.

\begin{table*}[t!]
\captionsetup{justification=centering, labelsep=newline}
\caption{Performance evaluation of state-of-the-art methods and the proposed network.}
    \centering
    \begin{tabularx}{\textwidth}{m{4.5cm}|c|>{\centering\arraybackslash}X>{\centering\arraybackslash}X>{\centering\arraybackslash}X|>{\centering\arraybackslash}X|>{\centering\arraybackslash}X}
        Methods & Backbone & AP & $\text{AP}_{50}$ & $\text{AP}_{75}$ & mIoU & FPS\\
        \hline
        Faster R-CNN \cite{ren2015faster} (1 class) & ResNet-18 & 0.69 & 0.88 & 0.51 & 0.78 & 9.02\\
        Faster R-CNN \cite{ren2015faster} (32 classes) & ResNet-18 & 0.64 & 0.81 & 0.50 & 0.76 & 5.68\\
        \hline
        \multirow{3}{*}{CenterNet \cite{duan2019centernet} (1 class)} & ResNet-18 & 0.60 & 0.81 & 0.17 & 0.57 & \textbf{35.38}\\
            & DLA-34 & 0.69 & \textbf{0.91} & 0.42 & 0.66 & \textbf{17.40}\\
            & HG-Stacked & 0.68 & 0.90 & 0.43 & 0.66 & \textbf{9.38}\\
        \hline
        \multirow{3}{*}{CenterNet \cite{duan2019centernet} (32 classes)} & ResNet-18 & 0.67 & 0.90 & 0.34 & 0.70 & 34.83\\
            & DLA-34 & 0.55 & 0.72 & 0.12 & 0.63 & 16.84\\
            & HG-Stacked & 0.70 & \textbf{0.91} & 0.54 & 0.75 & 9.17\\
        \hline
        \multirow{3}{*}{Our method} & ResNet-18 & \textbf{0.81} & \textbf{0.91} & \textbf{0.90} & \textbf{0.84} & 18.58\\
            & DLA-34 & \textbf{0.80} & \textbf{0.91} & \textbf{0.81} & \textbf{0.84} & 12.91\\
            & HG-Stacked & \textbf{0.77} & \textbf{0.91} & \textbf{0.80} & \textbf{0.81} & 7.88\\
    \end{tabularx}
    \label{table:accuracy}
\end{table*}

\begin{table*}[t!]
\captionsetup{justification=centering, labelsep=newline}
\caption{Performance on ablations of the proposed network.}
    \centering
    \begin{tabularx}{\textwidth}{m{4.5cm}|c|>{\centering\arraybackslash}X>{\centering\arraybackslash}X>{\centering\arraybackslash}X|>{\centering\arraybackslash}X}
        Methods & Backbone & AP & $\text{AP}_{50}$ & $\text{AP}_{75}$ & mIoU\\
        \hline
        \multirow{3}{*}{Our method w/o DR} & ResNet-18 & \textbf{0.79} & \textbf{0.91} & \textbf{0.81} & \textbf{0.82}\\
            & DLA-34 & \textbf{0.80} & \textbf{0.91} & \textbf{0.80} & \textbf{0.83}\\
            & HG-Stacked & \textbf{0.76} & \textbf{0.91} & \textbf{0.80} & \textbf{0.79}\\
        \hline
        \multirow{3}{*}{Our method w/o OFF} & ResNet-18 & 0.58 & 0.69 & 0.23 & 0.60\\
            & DLA-34 & 0.58 & 0.69 & 0.24 & 0.61\\
            & HG-Stacked & 0.63 & 0.80 & 0.33 & 0.67\\
        \hline
        \multirow{3}{*}{Our method w/o DR and OFF} & ResNet-18 & 0.57 & 0.69 & 0.22 & 0.60\\
            & DLA-34 & 0.58 & 0.69 & 0.24 & 0.60\\
            & HG-Stacked & 0.63 & 0.81 & 0.32 & 0.67\\
    \end{tabularx}
    \label{table:accuracy_ablations}
\end{table*}

\subsection{Evaluation Metrics}
To measure the accuracy of tooth detection, we employed average precision (AP) \cite{everingham2010pascal} metric. This metric is calculated based on the area under the receiver operating characteristics (ROC) curve by considering the criterion of intersection-over-union (IoU) \cite{everingham2010pascal}:
\begin{equation}
    IoU=\frac{B_{inf}\cap B_{gt}}{B_{inf}\cup B_{gt}},
\label{eq:iou}
\end{equation}
where $B_{inf}$ is the inferred box area and $B_{gt}$ is the corresponding ground-truth box area. The precision (i.e., $\frac{TP}{TP+FP}$) and recall (i.e., $\frac{TP}{TP+FN}$) can be calculated, where TP, FP, and FN indicate the number of true positives, false positives, and false negatives, respectively, according to a certain IoU threshold value. We calculated the area under the ROC curve based on 0.05 interval of the IoU threshold ranging from 0 to 1.\par

To measure the accuracy of localization in successful detection, we introduced the mean IoU (mIoU) metric. We first matched the similarities between the detected boxes and the ground-truth boxes based on the criterion of maximum IoU. The detected boxes were assigned to a single ground-truth box once its IoU value was bigger than zero. For the final assignment, the ground-truth boxes selected a single detected box that had themaximum IoU value among the assigned. The mIoU value is calculated based on the matched pair of boxes without false detection:
\begin{equation}
    mIoU=\frac{\Sigma_{i=1}^N IoU_i}{N},
\label{eq:miou}
\end{equation}
where $N$ is the number of matched pairs.\par

To measure the identification accuracy, we measured the precision and recall metrics:
\begin{equation}
    Precision=\frac{N_{TPN}}{N_{DB}}
\label{eq:nprec}
\end{equation}
and
\begin{equation}
    Recall=\frac{N_{TPN}}{N_{GTB}},
\label{eq:nrec}
\end{equation}
where $N_{GTB}$, $N_{DB}$, and $N_{TPN}$ denote the number of ground-truth tooth boxes (i.e., existing teeth), detected boxes with respect to a threshold value of $IoU=50$, and true positive identifications (i.e., accurate numbering) among $N_{DB}$, respectively.

\subsection{Accuracy Evaluation}
We performed a comparative analysis of the tooth detection results with state-of-the-art detectors, including anchor-based faster-RCNN \cite{ren2015faster} and center-point-based method \cite{duan2019centernet}. We present two different metrics: tooth or non-tooth classification (i.e., 1 class) and classification of all 32 teeth (i.e., 32 classes) for each network. Various backbone networks including ResNet-18 \cite{he2016deep}, DLA-34 \cite{yu2018deep}, HG-Stacked \cite{newell2016stacked} were used in the experiment for a comprehensive analysis, as presented in \cite{duan2019centernet}. Table \ref{table:accuracy} lists the values of AP, mIoU, and the computational performance based on frames per second (FPS). The accuracy of detection was evaluated by considering several AP values, i.e., AP \cite{everingham2010pascal}, $\text{AP}_{50}$, and $\text{AP}_{75}$. $\text{AP}_{50}$ and $\text{AP}_{75}$ indicate the F1 scores where the IoU threshold is given by 50 and 75, respectively. A localization accuracy of successful detection was evaluated by mIoU, as presented in (\ref{eq:miou}).\par

The results demonstrate that our proposed method is superior to other state-of-the-art methods pertaining to tooth detection. Regardless of the backbone networks, the proposed method showed the best AP and mIoU scores when compared to other detection approaches. The similar AP scores of our method for each backbone network indicate that the complexity of the network (i.e., network architecture) is not an important factor for the final accuracy. Among the networks, CenterNet \cite{duan2019centernet} showed the highest FPS.\par

\begin{figure*}[tb]
    \centering
        \begin{minipage}[b]{1.33in}
            \captionsetup[subfigure]{labelformat=parens,labelsep=space,font=small}
            \centering
                \vfil
                \includegraphics[width=1.3in]{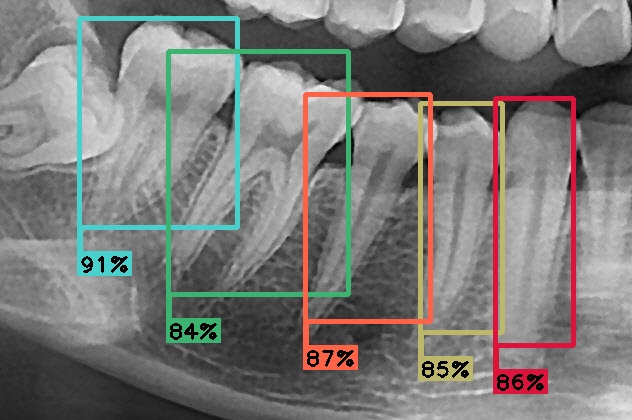}
                \includegraphics[width=1.3in]{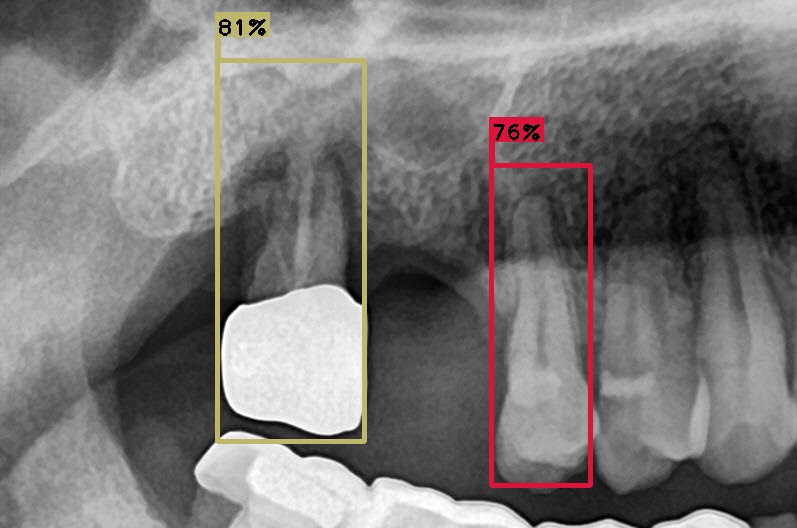}
                \includegraphics[width=1.3in]{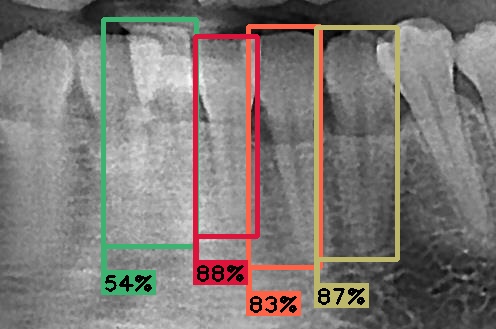}
            \captionof{subfigure}{Faster R-CNN-1}
        \end{minipage}
        \hfil
        \begin{minipage}[b]{1.33in}
            \captionsetup[subfigure]{labelformat=parens,labelsep=space,font=small}
            \centering
                \vfil
                \includegraphics[width=1.3in]{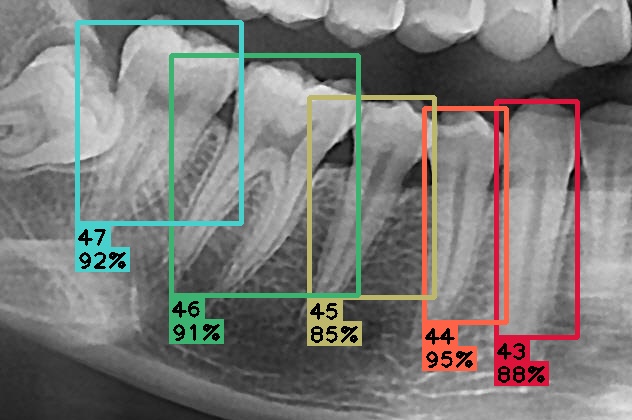}
                \includegraphics[width=1.3in]{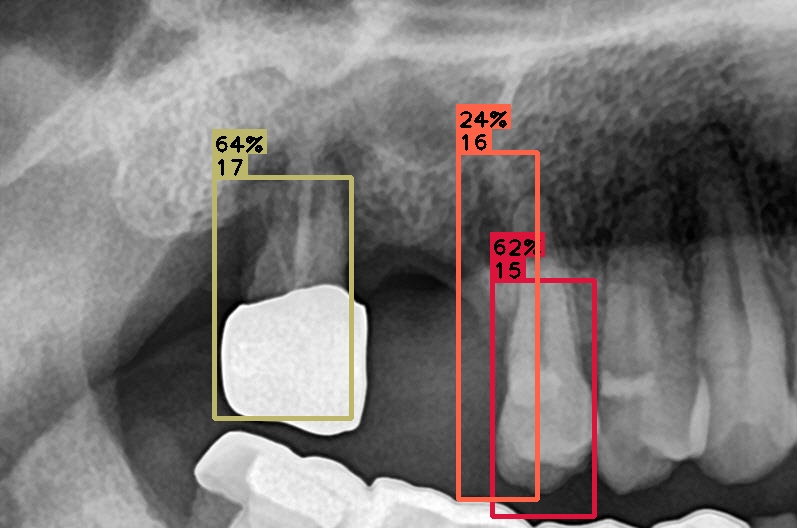}
                \includegraphics[width=1.3in]{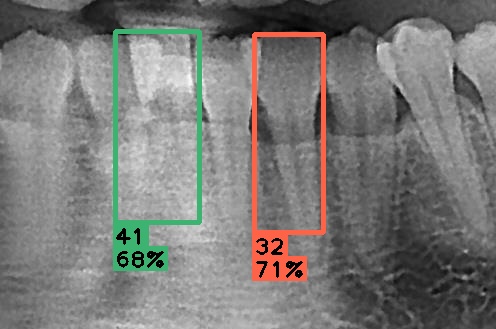}
            \captionof{subfigure}{Faster R-CNN-32}
        \end{minipage}
        \hfil
        \begin{minipage}[b]{1.33in}
            \captionsetup[subfigure]{labelformat=parens,labelsep=space,font=small}
            \centering
                \vfil
                \includegraphics[width=1.3in]{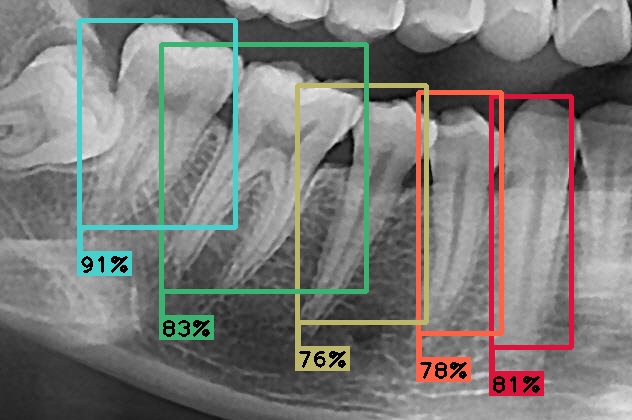}
                \includegraphics[width=1.3in]{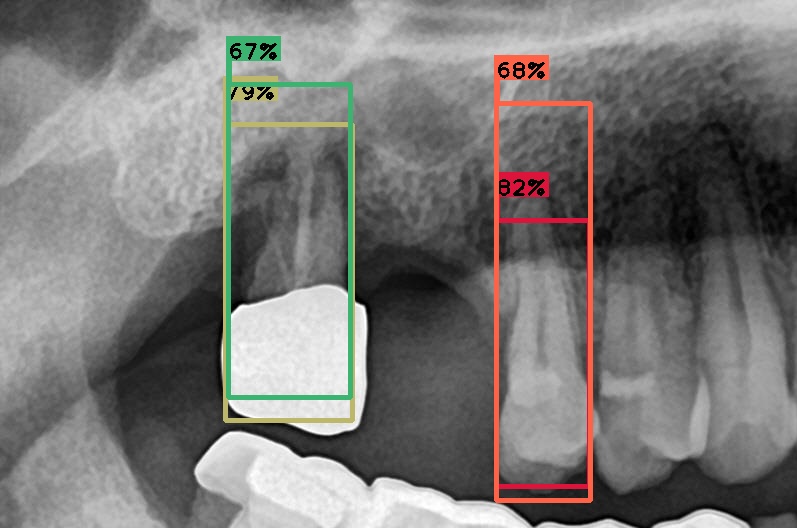}
                \includegraphics[width=1.3in]{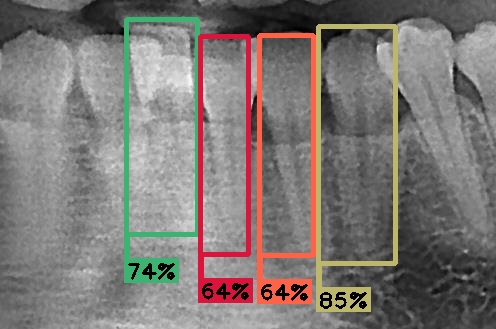}
            \captionof{subfigure}{CenterNet-1}
        \end{minipage}
        \hfil
        \begin{minipage}[b]{1.33in}
            \captionsetup[subfigure]{labelformat=parens,labelsep=space,font=small}
            \centering
                \vfil
                \includegraphics[width=1.3in]{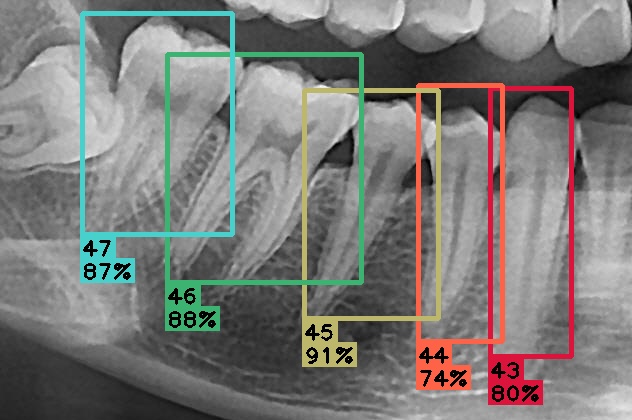}
                \includegraphics[width=1.3in]{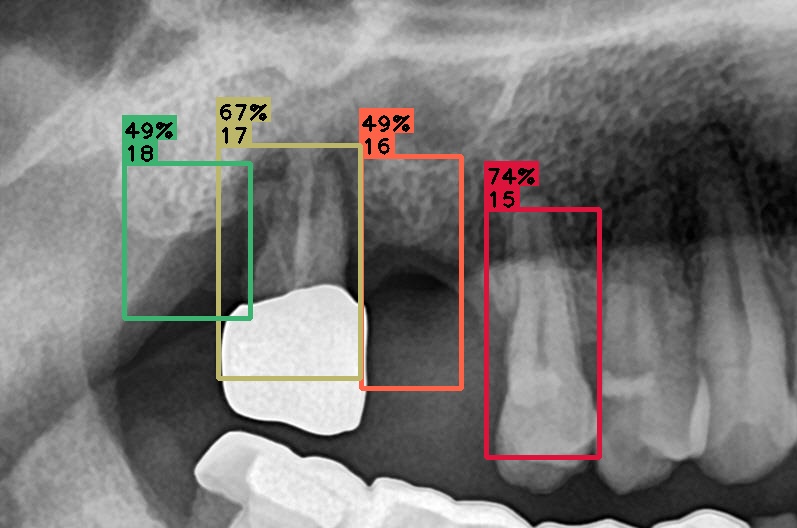}
                \includegraphics[width=1.3in]{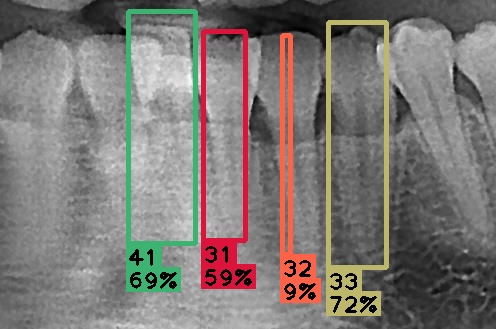}
            \captionof{subfigure}{CenterNet-32}
        \end{minipage}
        \hfil
        \begin{minipage}[b]{1.33in}
            \captionsetup[subfigure]{labelformat=parens,labelsep=space,font=small}
            \centering
                \vfil
                \includegraphics[width=1.3in]{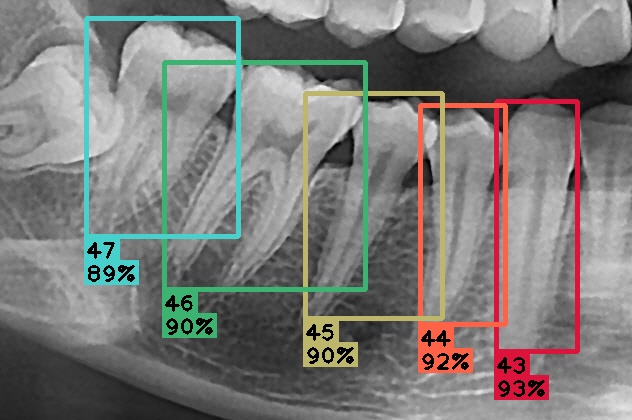}
                \includegraphics[width=1.3in]{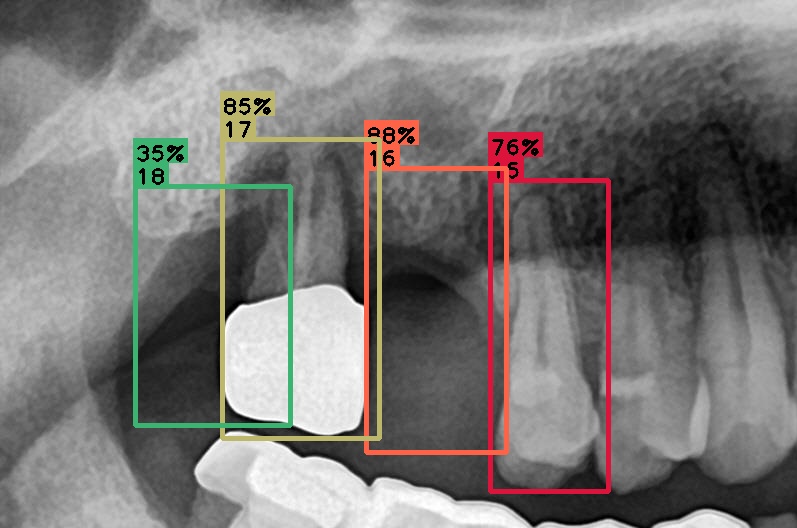}
                \includegraphics[width=1.3in]{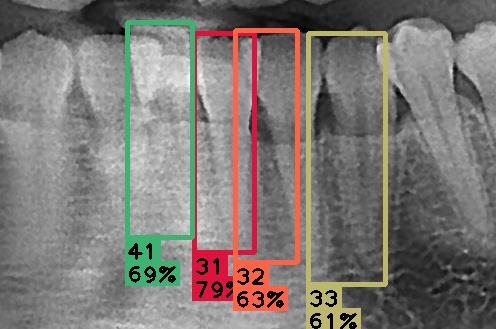}
            \captionof{subfigure}{Our method}
        \end{minipage}
        \vfil

    \caption{Visualization of detection results.}
    \label{fig:vis}
\end{figure*}

Figure \ref{fig:vis} illustrates certain sample results. The numbers denote the anatomical tooth identifiers and the percentage values indicate the IoU values. The proposed method accurately localized each individual tooth when compared to other state-of-the-art methods. The networks that trained with one class, i.e., faster R-CNN-1 and CenterNet-1, did not detect non-teeth regions because the networks only detected existing teeth from the input images. The proposed method demonstrated the most successful results among the other networks regardless of the missing teeth (the second row in Fig. \ref{fig:vis}).

\subsection{Ablation Studies}
To verify the effect of our proposed method, several ablation studies were conducted (Table \ref{table:accuracy_ablations}), i.e., without DR (i.e., (\ref{eq:dr})), without offset (OFF; i.e., (\ref{eq:offset})), and without both DR and OFF. As listed in Table \ref{table:accuracy_ablations}, the improvement of accuracy was primarily achieved through the multitask offset training. The accuracy of detection severely decreased without the offset branch (i.e., our proposed method without OFF). The distance-based regularization loss showed no significant difference without the offset training. The best accuracy was obtained by employing both DR and OFF losses (Table \ref{table:accuracy}).\par

\begin{table}[t!]
\captionsetup{justification=centering, labelsep=newline}
\caption{Center point regression errors of the proposed network.}
    \centering
    \begin{tabularx}{\linewidth}{c|c|>{\centering\arraybackslash}X>{\centering\arraybackslash}X}
        Methods & Backbone & ${\text{MSE}}_1$ & ${\text{MSE}}_2$\\
        \hline
        \multirow{3}{*}{Our method} & ResNet-18 & 207.00 & \textbf{23.98}\\
            & DLA-34 & 204.86 & \textbf{22.06}\\
            & HG-Stacked & 121.01 & \textbf{22.72}\\
        \hline
        \multirow{3}{*}{\makecell{Our method \\w/o DR}} & ResNet-18 & 248.63 & 32.61\\
            & DLA-34 & 207.40 & 24.46\\
            & HG-Stacked & 119.21 & 25.81\\
        \hline
        \multirow{3}{*}{\makecell{Our method \\w/o OFF}} & ResNet-18 & \textbf{200.22} & N/A\\
            & DLA-34 & \textbf{191.02} & N/A\\
            & HG-Stacked & 110.29 & N/A\\
        \hline
        \multirow{3}{*}{\makecell{Our method \\w/o DR and OFF}} & ResNet-18 & 211.78 & N/A\\
            & DLA-34 & 199.47 & N/A\\
            & HG-Stacked & \textbf{108.16} & N/A\\
    \end{tabularx}
    \label{table:accuracy_mse}
\end{table}

\begin{figure}[t!]
    \centering
    \subfloat[Ground-truth annotated points]{\includegraphics[width=\linewidth]{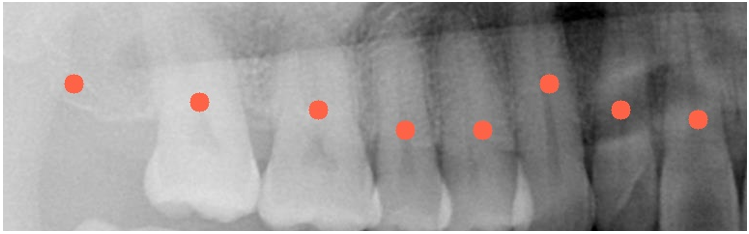}%
    \label{fig:vis_center_gt}}
    \vfil
    \subfloat[Our method]{\includegraphics[width=\linewidth]{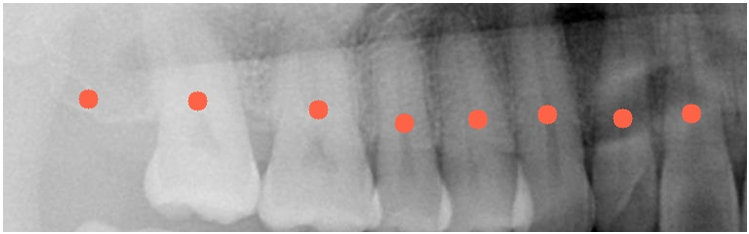}%
    \label{fig:vis_center_ours}}
    \vfil
    \subfloat[Our method without DR loss]{\includegraphics[width=\linewidth]{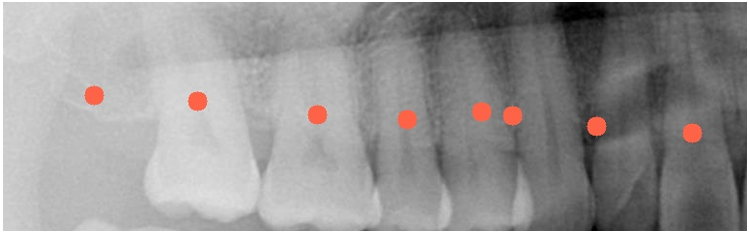}%
    \label{fig:vis_center_ours-dr}}
    \caption{Visualization of center points: (a) Ground-truth annotated points; (b) Our method; (c) Our method without DR loss. Distances between neighboring teeth are regularized by employing DR loss (b) compared to (c).}
    \label{fig:vis_center}
\end{figure}

The MSE values for center point regression are also listed in Table \ref{table:accuracy_mse} to assess the performance of localization. ${\text{MSE}}_1$ denotes the MSE loss in the first estimation of center points (i.e., \textit{center points} in Fig. \ref{fig:network}). Similarly, ${\text{MSE}}_2$ denotes the MSE loss for the final center point regression calculated from the first estimation and the offset (i.e., \textit{center offsets} in Fig. \ref{fig:network}). The accuracy of center point regression significantly improved owing to multitask offset training (i.e., ${\text{MSE}}_2$). The proposed DR loss also improved the accuracy of the first and final outputs. In the case of the network without offset, the first accuracy improved, which indicates that the proposed cascaded fashion of two-step training concentrated on the final accuracy rather than the first regression. Sample visualizations of center points regression are illustrated in Fig. \ref{fig:vis_center}. The proposed DR successfully regularized the proximate distances between neighboring teeth; thus, an improvement in regression accuracy was observed.

\begin{figure*}[t!]
    \centering
    \subfloat[Our proposed network]{\includegraphics[width=2in]{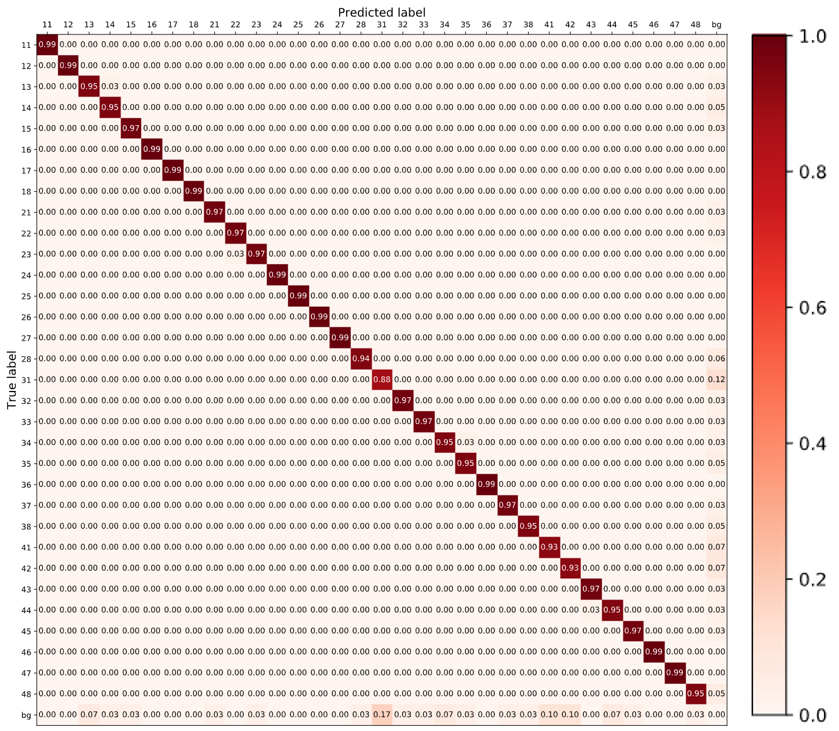}%
    \label{fig:confusion_1}}
    \hfil
    \subfloat[Faster R-CNN-32 \cite{ren2015faster}]{\includegraphics[width=2in]{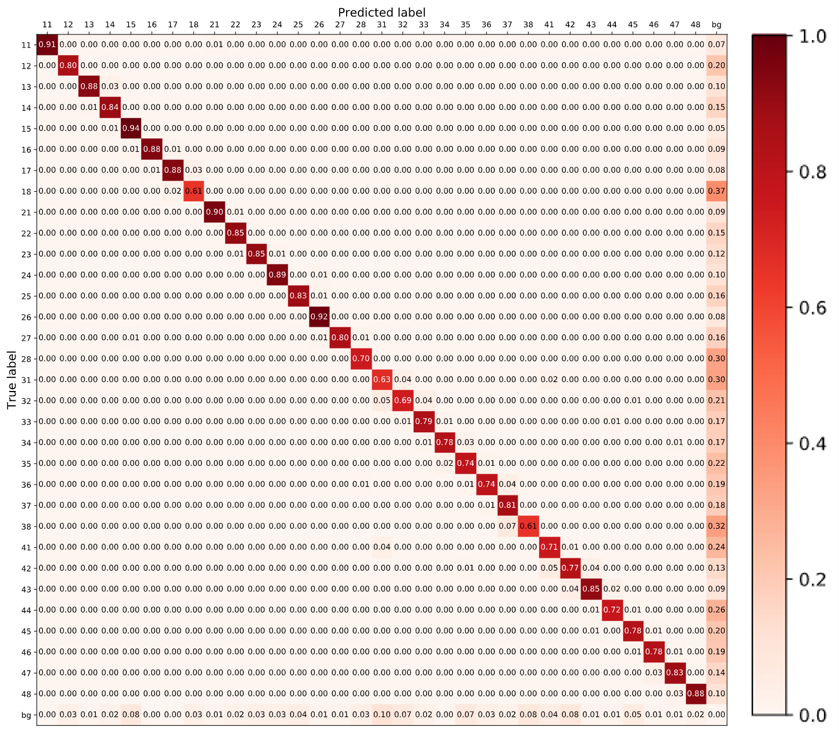}%
    \label{fig:confusion_2}}
    \hfil
    \subfloat[CenterNet-32 \cite{duan2019centernet}]{\includegraphics[width=2in]{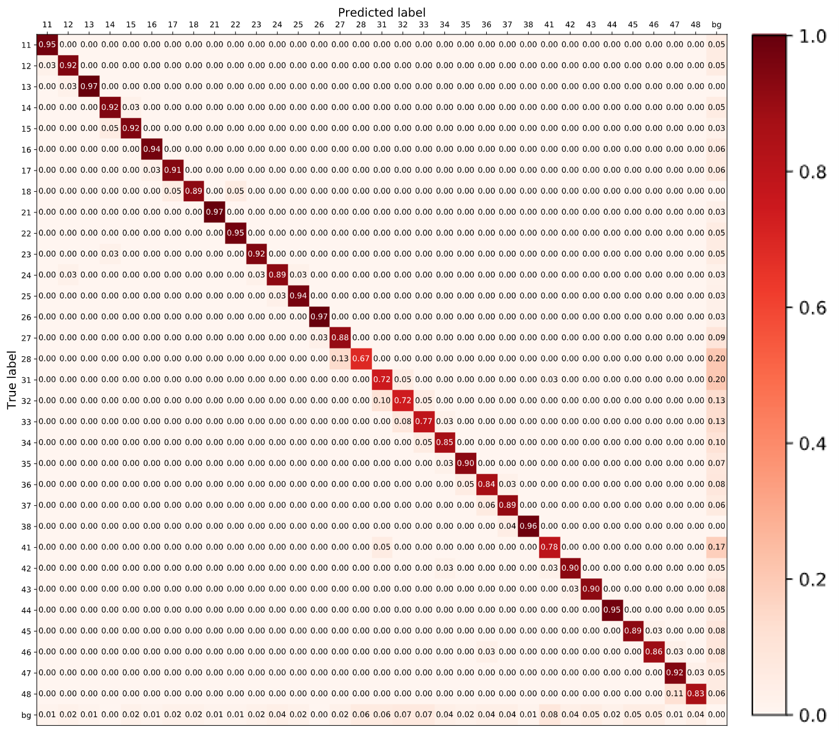}%
    \label{fig:confusion_3}}
    \caption{Confusion matrix examples based on ResNet-18 \cite{he2016deep} backbone network. Colors are mapped for the normalized confusion matrices.}
    \label{fig:confusion}
\end{figure*}

\begin{table}[t!]
\captionsetup{justification=centering, labelsep=newline}
\caption{Performance of tooth identification.}
    \centering
    \begin{tabularx}{\linewidth}{c|c|>{\centering\arraybackslash}X>{\centering\arraybackslash}X}
        Methods & Backbone & Precision & Recall\\
        \hline
        \makecell{Faster R-CNN-32 \\\cite{ren2015faster}} & ResNet-18 & 0.961 & 0.805\\
        \hline
        \multirow{3}{*}{\makecell{CenterNet-32 \\\cite{duan2019centernet}}} & ResNet-18 & 0.958 & 0.889\\
            & DLA-34 & 0.982 & 0.792\\
            & HG-Stacked & 0.967 & 0.928\\
        \hline
        \multirow{3}{*}{Our method} & ResNet-18 & \textbf{0.997} & \textbf{0.972}\\
            & DLA-34 & \textbf{0.997} & \textbf{0.971}\\
            & HG-Stacked & 0.986 & \textbf{0.960}\\
        \hline
        \multirow{3}{*}{\makecell{Our method \\w/o DR}} & ResNet-18 & 0.973 & 0.931\\
            & DLA-34 & 0.986 & 0.962\\
            & HG-Stacked & 0.982 & 0.957\\
        \hline
        \multirow{3}{*}{\makecell{Our method \\w/o OFF}} & ResNet-18 & 0.994 & 0.721\\
            & DLA-34 & 0.979 & 0.719\\
            & HG-Stacked & \textbf{0.992} & 0.843\\
        \hline
        \multirow{3}{*}{\makecell{Our method \\w/o DR and OFF}} & ResNet-18 & 0.993 & 0.720\\
            & DLA-34 & 0.993 & 0.720\\
            & HG-Stacked & 0.991 & 0.864\\
    \end{tabularx}
    \label{table:identification}
\end{table}

\subsection{Tooth Identification}
The accuracy of identification was evaluated by comparing the proposed method with faster R-CNN-32 \cite{ren2015faster} and CenterNet-32 \cite{duan2019centernet}. The network was trained to classify each tooth by its anatomical identifier based on the ground-truth tooth numbers. Table \ref{table:identification} lists the results of the precision (according to (\ref{eq:nprec})) and recall (according to (\ref{eq:nrec})) values. The accuracy was evaluated based on the existing teeth in the test images. Our proposed method showed superior performance in teeth identification. The primary advantage was obtained by the fixed 32-point regression and DR loss. Our method does not require any additional classifications as employed in other detection methods such as faster R-CNN-32 \cite{ren2015faster} and CenterNet-32 \cite{duan2019centernet}. Ablation cases of the proposed network demonstrated that our proposed DR and OFF losses significantly improved the accuracy of identification. The recall value significantly decreased without offset training. Figure \ref{fig:confusion} illustrates sample visualizations of the confusion matrix based on the ResNet-18 \cite{he2016deep} backbone network.


\section{Discussion}
Individual tooth detection and identification is a clinical CAD application that requires a high standard of accuracy for its deployment in clinics. It is challenging to obtain an accurate system by employing the state-of-the-art object detection methods \cite{ren2015faster, duan2019centernet} because there are no specific metrics to minimize false positive or false negative detection. In this study, we proposed a fixed 32-point regression method and simultaneous localization that resolves these limitations. The proposed method automatically identifies the anatomical identifiers of teeth rather than performing each classification as presented in other studies \cite{ren2015faster, duan2019centernet}. Thus, the proposed network achieved high accuracy in identification, which is important for the application. The proposed 32-point regression method is similar to the landmark detection approaches \cite{sun2013deep, zhang2014facial, toshev2014deeppose} that directly estimate the positions of fixed points. Our work combined the point regression scheme with object localization to achieve an accurate detection task with identification.\par

Instead of employing a single feed-forward network to detect 32 different classes, which contains explicit classification steps, we employed a cascaded framework to implement class-agnostic object localization. The proposed architecture is more efficient than other state-of-the-art methods because there are no additional classification layers for each class.

\section{Conclusion}
In this study, we proposed a point-wise tooth localization neural network by introducing a spatial DR loss. The proposed network employed center point regression for all the anatomical teeth (i.e., 32 points), which automatically identified each tooth. The $L_2$ regularization loss for Laplacian of spatial distances improved the detection accuracy based on center points. The final detection was performed using a cascaded, class-agnostic localization neural network with multitask training of center offsets. The experimental results demonstrated that the proposed method outperformed state-of-the-art detection approaches, which typically require an external classification for each class. Our proposed method achieved a precision of 0.997 and recall value of 0.972 for tooth identification, indicating the practical applicability of the proposed system in clinics.

\section*{Acknowledgement}
This work was partly supported by Institute for Information \& communications Technology Promotion(IITP) grant funded by the Korea government(MSIT) (No.2017-0-01815, Development of AR-based Surgery Toolkit and Applications). And this research was partly supported by the Basic Science Research Program through the National Research Foundation of Korea (NRF) funded by the Ministry of Science, ICT and Future Planning (No. 2017R1D1A1B03034484). And this research was partly supported by the Basic Science Research Program through the National Research Foundation of Korea (NRF) funded by the Ministry of Science, ICT and Future Planning (No. 2017R1A2B3011475).

{\small
\bibliographystyle{IEEEtran}

\bibliography{mybib}
}

\end{document}